\documentclass[letterpaper, 10 pt, conference]{ieeeconf}  

\IEEEoverridecommandlockouts                              

\usepackage{multicol}
\usepackage[bookmarks=true]{hyperref}
\usepackage{booktabs}
\usepackage{cite}
\usepackage[ruled,vlined,linesnumbered]{algorithm2e}

\usepackage{amsmath}
\usepackage{amssymb}
\usepackage{xcolor}

\usepackage{graphicx}
\usepackage{subcaption}
\usepackage{multirow}

\usepackage{tikz}
\usetikzlibrary{arrows.meta,calc}

\title{\LARGE \bf
Manip4Care: Robotic Manipulation of Human Limbs for Solving Assistive Tasks
}

\author{Yubin Koh and Ahmed H. Qureshi
\thanks{The authors are with the Department of Computer Science, Purdue University, 
        West Lafayette, IN, USA
        {\tt\small \{koh22, ahqureshi\}@purdue.edu}}%
}

\begin{document}

\maketitle
\thispagestyle{empty}
\pagestyle{empty}

\begin{abstract}
Enabling robots to grasp and reposition human limbs can significantly enhance their ability to provide assistive care to individuals with severe mobility impairments, particularly in tasks such as robot-assisted bed bathing and dressing. However, existing assistive robotics solutions often assume that the human remains static or quasi-static, limiting their effectiveness. To address this issue, we present Manip4Care, a modular simulation pipeline that enables robotic manipulators to grasp and reposition human limbs effectively. Our approach features a physics simulator equipped with built-in techniques for grasping and repositioning while considering biomechanical and collision avoidance constraints. Our grasping method employs antipodal sampling with force closure to grasp limbs, and our repositioning system utilizes the Model Predictive Path Integral (MPPI) and vector-field-based control method to generate motion trajectories under collision avoidance and biomechanical constraints. We evaluate this approach across various limb manipulation tasks in both supine and sitting positions and compare outcomes for different age groups with differing shoulder joint limits. Additionally, we demonstrate our approach for limb manipulation using a real-world mannequin and further showcase its effectiveness in bed bathing tasks.
Our implementation is available at \url{https://github.com/yubink2/Manip4Care}.

\end{abstract}






\section{Introduction}

Robot-assisted care offers the promise of mitigating the caregiver burden and improving the quality of life for individuals with severe mobility impairments \cite{christensen2021roadmap}. 
In particular, tasks such as bed bathing and dressing often require repositioning a person’s limb to reach difficult areas or align garments correctly. Beyond providing assistance, addressing limb manipulation challenges is essential for achieving long-term goals in robot-assisted physiotherapy. Moreover, limb manipulation is also important for retrieving individuals during search and rescue missions. 
Despite this clear need, many existing assistive robots treat the human as entirely passive—keeping limb posture fixed throughout the task—which severely limits their ability to provide comprehensive care and long-term assistance.

Most research on robotic manipulation of the human body has centered on rehabilitation devices (e.g., wheelchairs, exoskeletons) \cite{mohebbi2020human, sivakanthan2021person}, with only a few studies formulating human‐body manipulation as a constrained optimization problem \cite{yip2023manipulation}. However, these efforts primarily focus on lifting or repositioning an entire body, which is less appropriate for localized limb manipulation, where the scope of the problem can be simplified. Moreover, it remains difficult to reproduce or generalize solutions across multiple tasks, largely due to a lack of open-source frameworks that accurately represent human limb biomechanics and the complexities of close-proximity physical robot–human interaction. To address this gap, our work offers a pipeline for general‐purpose limb manipulation between any initial and goal limb poses.

In this paper, we propose Manip4Care, an open-source, modular simulation pipeline for flexible robotic limb manipulation in assistive care.
In addition, we present a first approach that solves the bed bathing tasks via limb manipulation. Our key contributions are listed as follows:
\begin{itemize}
\item \textbf{Antipodal Sampling-based Limb Grasping} that ensures force closure and selects the grasp for effective limb manipulation.

\item \textbf{MPPI-based Trajectory Planning} that repositions limbs between configurations under biomechanical and collision constraints.

\item \textbf{Vector Field-based Trajectory Follower} that guides the robot to follow a specified trajectory and avoid collisions and violations of biomechanical constraints.

\item \textbf{Reactive Path Generation and Following} that iteratively updates trajectories to safely guide the limb to its target pose.

\item \textbf{Manip4Care and a Bed Bathing Demonstration} that iteratively repositions limbs to maximize wiping coverage, supported by our open-source simulation setup for assistive applications.
 
\end{itemize}

 
\section{Related Work}
This section outlines our relevant work. To the best of our knowledge, no existing frameworks specifically address general-purpose limb repositioning using robotic manipulators. Therefore, we will highlight previous research in related fields that touch upon the task of limb manipulation. Additionally, we will discuss the related work concerning the application of solving bed bathing tasks.

Existing approaches to limb repositioning tasks rely on exoskeletons \cite{mohebbi2020human, 2017upperrehab, 2017rehab, 2018rehabilitation} or human-operated robot systems \cite{sivakanthan2021person, mukai2010development, onishi2007generation}. 
Exoskeletons are primarily intended to enhance human mobility, whereas our focus is to provide human with personal care and assistance in daily life tasks such as bathing and dressing. In this context, user-centered design studies emphasize adaptable, safe solutions in personal care settings \cite{2018hrcollab, 2017repositionsurvey}. 
On the other hand, human-operated robots rely on skilled human operators to perform limb manipulation tasks. Planning to maneuver a human body in a search-and-rescue scenario has also been explored by employing biomechanical constraints to minimize injury risk \cite{yip2023manipulation}. However, many of these methods are not directly applicable for localized limb repositioning. A more closely related approach is limb repositioning via model predictive control method \cite{kemp2016reposition}, but it focuses on short-horizon force control rather than a comprehensive motion-planning framework. To the best of our knowledge, our approach is the first general-purpose limb manipulation method that enables versatile repositioning for a wide range of assistive tasks. 

In the remainder of this section, we highlight the importance of limb manipulation tasks for building assistive robots, which can improve the lives of individuals with limited mobility by offering support in daily tasks such as feeding, dressing, and personal hygiene. Simulation-based platforms have been created to benchmark robot controllers for a variety of these tasks \cite{erickson2020assistivegym}. Early demonstrations of bed-bathing robots employed soft, compliant end effectors to clean a person’s arm, showing that gentle and effective personal hygiene is feasible \cite{kemp2010bedbath}. Recent efforts integrate compliance and advanced multimodal perception \cite{2024rabbit}, as well as visuo-tactile feedback \cite{Gu2024VTTBAV}, to enhance sensing accuracy and reduce contact forces during wiping. However, these systems assume a static human limb configuration, restricting the cleaning process to a single arm segment at a time and often requiring manual repositioning to achieve complete coverage. Therefore, incorporating robust limb manipulation techniques to solve these bed bathing tasks is critical for facilitating people with mobility impairments. 

\section{Methods}
\begin{figure}
    \centering
    \includegraphics[width=0.99\linewidth, trim=0 5mm 0 5mm, clip]{"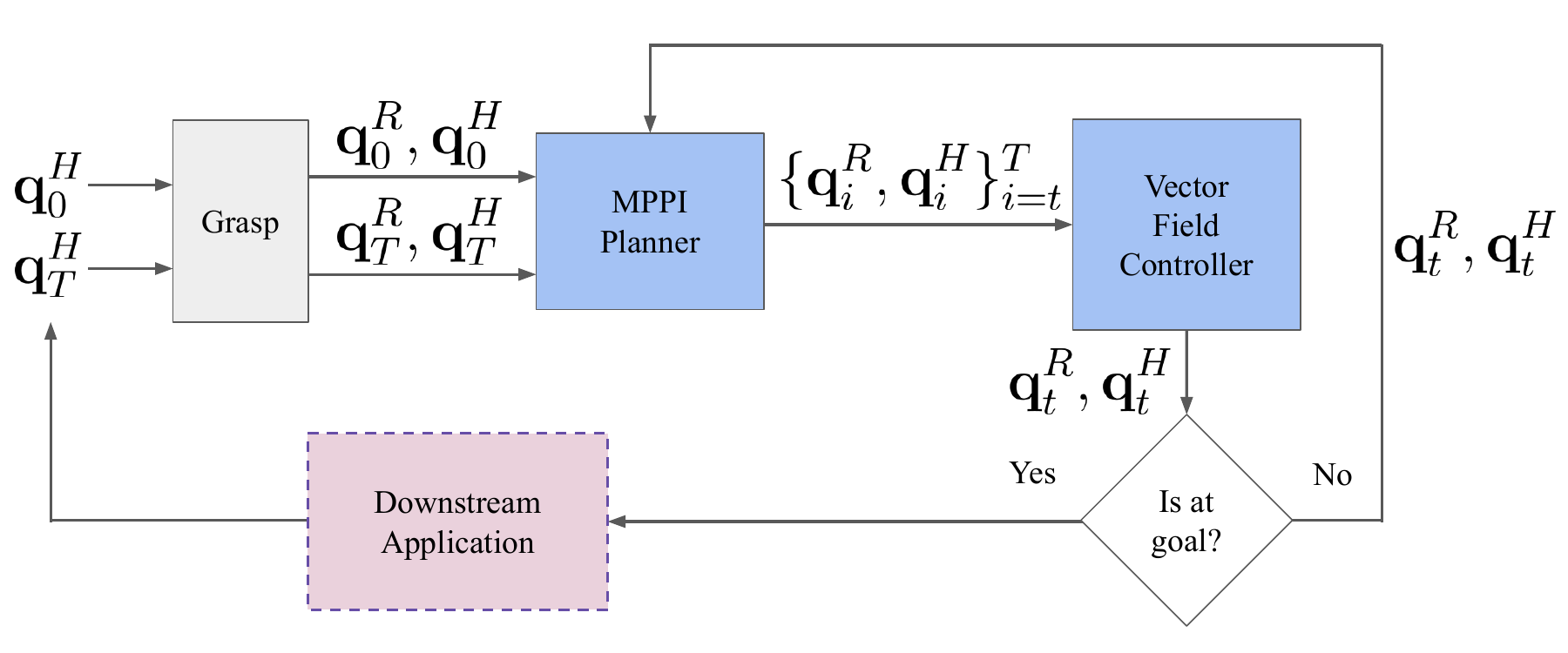"}
    \caption{Overview of the complete limb manipulation pipeline and its integration with the downstream applications such as bed bathing and assistive dressing.}
    \vspace{-3mm}
    \label{fig:pipeline}
\end{figure}
In this section, we outline our methodology for manipulating human limbs and its integration with downstream applications such as bed bathing as illustrated in Fig.~\ref{fig:pipeline}.
The downstream application provides the target limb configuration, and our system grasps and manipulates the human limb from its initial configuration to the target configuration adhering to biomechanical and collision avoidance constraints. 
The process repeats in the loop until the objectives of downstream tasks are accomplished.


\subsection{Notations and Problem Statement}
\label{sec:problem}
This section presents our necessary notations, assumptions, and formulation of human-robot close-loop biomechanical constraints and limb manipulation problems.

Throughout this paper, we use $\textbf{T}_a^b$ to denote the pose of $b$ in the coordinate frame of $a$, with $w$ as the world frame (e.g., $\textbf{T}_a^w$ denotes the pose of $a$ in the world frame). We use $\textbf{p}$ and $\textbf{R}$ to represent the position and rotation of a given rigid body in the world frame. Superscripts $R$ and $H$ denote whether a variable, set, vector, or matrix is associated with the robot or human, respectively.
Let $\textbf{q}^R \in \mathbb{R}^n$ be the joint angles of a robot with $n$ degrees of freedom and $\textbf{q}^H \in \mathbb{R}^m$ be the joint angles of a human limb with $m$ degrees of freedom. Let $\textbf{q}_0$ and $\textbf{q}_T$ denote the initial and goal joint angles, respectively, with $T$ denoting the time horizon for executing the limb manipulation. Let $Q^R$ and $Q^H$ be the robot and human configuration space, respectively, where $Q^H$ is subject to the feasible range of human limb joint angles collected from~\cite{ghang2023elbowrange, gates2016upperlimbrange}. 
Then, the objective of our limb manipulation method is as follows: to move the human limb from an initial configuration $\textbf{q}^H_0$ to a goal configuration $\textbf{q}^H_T$, find a trajectory $\textbf{q}^R_{0:T}$ such that $\forall t \in \{0, \dotsc, T\}$, (1) $\textbf{q}^R_t \in Q^R$, (2) $\textbf{q}^H_t \in Q^H$, where $\textbf{q}^H_t$ is derived from $\textbf{q}^R_t$ via forward and inverse kinematics, and (3) $\textbf{q}^R_t$ and $\textbf{q}^H_t$ are collision-free. Note that to solve the limb manipulation problem, the robot must also solve the grasping problem to achieve the given task. 

Once the robot end effector grasps the human limb, we assume that the grasp point remains fixed relative to both the robot’s end effector and the human limb. That is, we assume a rigid, non-slip grasp. In this setup, the \textit{contact point} is the precise location on the human limb at which the robot’s end effector establishes its grasp.
From a kinematic perspective, the human limb and the robot can be viewed as two separate kinematic chains, each with its own fixed base. Representing the limb as having a fixed base is a reasonable approximation, as stated in~\cite{kemp2016reposition}. Once a rigid contact is established, the terminal node of the robot’s chain (the end effector) is linked to an intermediate node in the human limb’s chain (the contact point). By defining a fixed transform at that contact point, the two chains become a single composite structure with a closed‐loop constraint. As a result, any movement of the robot’s end effector will impose corresponding motion on the human limb.

\begin{figure}
     \centering
     \begin{subfigure}[b]{0.4\columnwidth}
         \centering
         \includegraphics[width=\linewidth]{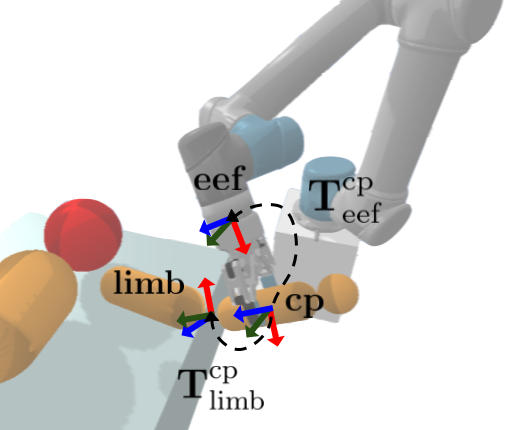}
         \caption{Supine Position}
     \end{subfigure}
     \hspace{0.3em}
     \begin{subfigure}[b]{0.4\columnwidth}
         \centering
         \includegraphics[width=\linewidth]{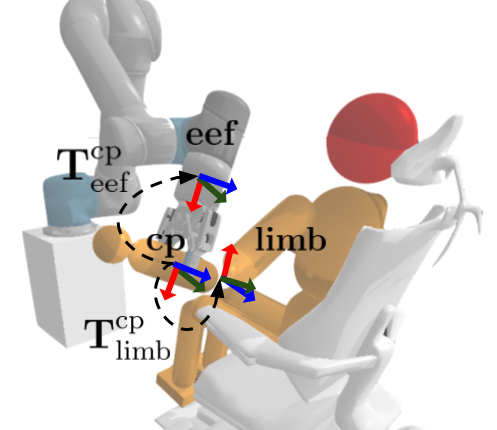}
         \caption{Sitting Position}
     \end{subfigure}
\caption{Example of the grasp selected on the forearm with a human in supine and sitting positions. 
Predefined transformation matrices for the chosen grasp are also illustrated.
}
\vspace{-3mm}
\label{fig:grasps}
\end{figure}

To represent these constraints, let $\textbf{T}_\text{eef}^\text{cp}$ denote the pose of the contact point in the coordinate frame of the robot's end effector. Let $\textbf{T}_\text{limb}^\text{cp}$ denote the pose of the contact point in the coordinate frame of the human limb. We assume that $\textbf{T}_\text{eef}^\text{cp}$ and $\textbf{T}_\text{limb}^\text{cp}$ remain fixed during limb manipulation. See Fig.~\ref{fig:grasps} for the visualization of the two transformation matrices.

\subsection{Grasp Generation}
Optimal grasp generation is essential to ensure safety and comfort during limb manipulation. To this end, we employ an antipodal sampling method~\cite{mahler2016dex} to generate grasp samples.
First, a point cloud of the limb to grasp onto is obtained and used to calculate surface normals for each point. 
Next, we identify pairs of points whose normals lie nearly opposite, which indicates that a parallel-jaw gripper placed at those points can apply opposing forces for a stable grasp.
We verify force closure by checking friction-cone overlap at both contact points. Each valid pair produces a candidate grasp. 

After a set of candidate grasps is generated, any non-reachable or in-collision grasps are discarded. Among the remaining $G$ candidates $\{g^R_i\}^G_{i=0}$, where each $g^R_i$ is a grasp pose associated with position $\textbf{p}^R_i \in \mathbb{R}^3$ and rotation $\textbf{R}^R_i \in \mathbb{R}^{3\times3}$, we choose a grasp that is (1) in a reasonable distance from the human joint to allow for moment generation and (2) oriented to be approximately perpendicular to the limb’s principal axis (e.g., from the shoulder to the elbow). 
This is to better align the end effector's rotational axes with the human limb.
We use the limb's pose in the world coordinate as a reference position $\textbf{p}^H_{ref}$ and a reference rotation $\textbf{R}^H_{ref}$. Then, we evaluate $\{g_i\}$ as follows:
\begin{equation}
\label{eq:grasp}
\begin{gathered}
    S_i = \alpha \cdot \,d_i- (1-\alpha) \cdot r_i,\:\:\: \text{with} 
    \\
    d_i = \bigl\| \textbf{p}^R_i - \textbf{p}^H_{ref} \bigl\|_2 ; 
    \quad r_i = \langle \textit{q}_i, \textit{q}_{ref} \rangle ;
    \\
    \textit{q}_i = \text{quat}(\textbf{R}^R_i) ; 
    \quad \textit{q}_{ref} = \text{quat}(\textbf{R}^R_{ref}) ;  
    \\
\end{gathered}
\end{equation}
where $\textit{q}$ is an unit quaternion converted from a rotation matrix and $\alpha \in [0, 1]$ balances out the two decision metrics.
In practice, we normalize $d_i$ and $r_i$ to make the score range between 0 and 1. Finally, a grasp $g^R_i$ with the smallest $S_i$ is selected for limb manipulation. Note that the final grasp will be kinematically reachable, free from collisions, aligned with the limb’s axis of rotation, and will maintain an adequate distance from the joints to allow for moment generation. Example grasps generated by our method for the human limb in a supine and sitting posture are shown in Fig. \ref{fig:grasps}.


\subsection{Limb Manipulation}
\label{sec:manipulation}
This section presents our limb manipulation approach.  The process begins with finding $\textbf{q}^R_0$ and $\textbf{q}^R_T$ corresponding to the initial human limb configuration $\textbf{q}^H_0$ and the desired human limb configuration $\textbf{q}^H_T$. This can be achieved using the predefined transformations $\textbf{T}_\text{eef}^\text{cp}$ and $\textbf{T}_\text{limb}^\text{cp}$ from Sec.~\ref{sec:problem} given the valid grasp associated with the initial limb configuration.
Then, the set of robot and human limb configuration pairs $\{\textbf{q}^R_0, \textbf{q}^H_0, \textbf{q}^R_T, \textbf{q}^H_T\}$ serves as inputs to our \emph{MPPI-Based Constrained Trajectory Planner}, which runs a hierarchical, model-predictive optimization to produce a collision‐free trajectory that respects both the human limb’s and the robot’s kinematic constraints. Finally, the resulting trajectory is tracked using a \emph{Vector Field Based Trajectory Follower}, allowing the system to smoothly perform limb repositioning and ensure continuous feasibility. We now discuss each of these components in detail.

\subsubsection{\textbf{MPPI-Based Constrained Trajectory Planner}}
Motion planning for articulated systems with strict biomechanical constraints demands more nuanced strategies than standard constrained motion planning problems, such as manipulating a cup. 
In those simpler tasks, the primary concern is to properly align the robot's end effector in a desired axis to interact with an external object. In contrast, when the robot is coupled to another articulated system (e.g., a human limb), the entire motion must consider \textit{both} the robot's joints \textit{and} the human limb's biomechanical constraints as part of one composite system. 
Since every incremental change in robot motion must respect the feasible range of the human limb, the feasible region for this composite human–robot system becomes significantly narrower and harder to explore.
To address these challenges, we implement a reactive, global trajectory planner based on Model Predictive Path Integral (MPPI)~\cite{2016mppi} that considers rigid coupling and feasible limb configuration space.
Our approach adapts MPPI from the Reactive Action and Motion Planner (RAMP)~\cite{2023ramp} framework, including its hierarchical structure for generating nominal control sequences. 
We extend RAMP's MPPI implementation from its original goal-reaching tasks under collision-avoidance constraints to handle limb manipulation tasks under closed-loop kinematic constraints.

We first begin with generating an initial hypothesis (nominal trajectory) that lies within both robot configuration space $Q^R$ and human limb configuration space $Q^H$. Given initial and goal human limb configurations $\textbf{q}^H_0$ and $\textbf{q}^H_T$, we compute an interpolated trajectory $\{\textbf{q}^H_0, \dotsc, \textbf{q}^H_T\}$. Then, for each $\textbf{q}^H_t$ in $t \in \{0, \dotsc, T\}$, we compute a corresponding robot configuration $\textbf{q}^R_t$ as follows.
\begin{gather*}
    \textbf{T}_\text{limb}^{w} = \text{FK}(\textbf{q}^H_t);
    \quad \textbf{T}_\text{eef}^{w} = \textbf{T}_\text{limb}^{w} \cdot \textbf{T}_\text{cp}^\text{limb} \cdot \textbf{T}_\text{eef}^\text{cp}; \\
    \textbf{q}^R_t = \text{IK}(\textbf{T}_\text{eef}^{w})
\end{gather*}

Once the initial hypothesis $\{\textbf{q}^R_0, \dotsc, \textbf{q}^R_T\}$ is established, we introduce stochastic perturbations around the nominal control sequence, sampling from a Gaussian distribution with temperature $\lambda=1$ and a covariance $\Sigma = 0.005 \cdot I$. 
We choose a small covariance to concentrate the search around the nominal control sequence, ensuring that sampled rollouts remain within the feasible human configuration space $Q^H$.
In addition, we employ a dynamic control horizon $H$ that scales with the length of the current trajectory hypothesis. That is, the control horizon is lengthened when the robot is far from its goal and shortened as it approaches it.
Then, we enforce the rigid pose constraints on each sampled control sequence using the \textit{IK handshaking} procedure \cite{berenson2011tsr} to ensure the robot’s end effector and the human limb remain properly aligned. This iterative approach refines each sampled rollout by first solving IK for the human limb based on the current end effector pose and clamping the human limb joint angles to ensure that they are within the feasible range. Then, we solve IK for the robot to match that updated human‐limb pose. Algorithm~\ref{alg:handshake} describes the complete procedure.

\begin{algorithm}[h]
\SetAlgoLined
\DontPrintSemicolon
\LinesNumbered
\SetKwComment{Comment}{/* }{ */}
\SetCommentSty{footnotesize}
\caption{IK-Handshaking ($\textbf{q}^R$, $\textbf{T}_\text{eef}^\text{cp}$, $\textbf{T}_\text{limb}^\text{cp}$)}
\label{alg:handshake}

    $ \textbf{T}_\text{eef}^\text{w} \gets \text{FK}_R(\textbf{q}^{R}) $ \;
    $ \textbf{T}_\text{limb}^\text{w} \gets \textbf{T}_\text{eef}^\text{w} \cdot \left(\textbf{T}_\text{cp}^\text{eef}\right)^{-1} \cdot \textbf{T}_\text{limb}^\text{cp} $ \;
    $ \textbf{q}^{H} \gets \text{IK}_H(\textbf{T}_\text{limb}^\text{w}) $ \;
    $ \textbf{q}^{H} \gets \text{CLAMP}_H(\textbf{q}^{H}) $ \Comment*[r]{To be within feasible range}
    $ \textbf{T}_\text{limb'}^\text{w} \gets \text{FK}_H(\textbf{q}^{H}) $ \Comment*[r]{Recompute limb pose}
    $ \textbf{T}_\text{eef'}^\text{w} \gets \textbf{T}_\text{limb'}^\text{w} \cdot \left(\textbf{T}_\text{limb}^\text{cp}\right)^{-1} \cdot \textbf{T}_\text{eef}^\text{cp}$ \Comment*[r]{New end effector pose}
    $ \textbf{q}^{R'} \gets \text{IK}_R(\textbf{T}_\text{eef'}^\text{w}) $  \; 
    \Return $\textbf{q}^{R'}$\; 

\end{algorithm}

Because each control sequence involves collision checks at multiple robot configurations, a rapid collision-checking method is essential. We achieve this by employing a Configuration Signed Distance Function (CSDF), which discretizes the robot and the human limb into sets of control points:
\[
\mathcal{P}^R = \{ \textbf{p}^R_i \mid i=1,\dotsc, N^R \} ; \ 
\mathcal{P}^H = \{ \textbf{p}^H_i \mid i=1,\dotsc, N^H \}
\]
where $N^R$ and $N^H$ are the numbers of control points for the robot and the human limb, respectively. 
First, we define a set of reference control points $\{ \textbf{p}^R_i \}$ for each robot link. At runtime, given a new robot configuration $\textbf{q}^R$, we apply GPU-accelerated forward kinematics to transform each reference point $\textbf{p}^R_i$ into its corresponding world-frame pose. Furthermore, we can compute $\{ \textbf{p}^H_i \}$ by transforming into the world frame through the same end‐effector pose. 
For every control point $\textbf{p}_i \in (\mathcal{P}^R \cup \mathcal{P}^H)$, the CSDF measures its signed distance to obstacles in the workspace, with a safety threshold $\rho = 2$ cm and an outer radius $r = 5$ for collision penalties.
Trajectory rollouts are then evaluated based on a cost function that considers (1) convergence to the goal configuration $\textbf{q}^R_T$, (2) collision avoidance, and (3) trajectory length. The best candidate is chosen as the proposed trajectory $\textbf{q}^R_{0:T}$ for execution.

\subsubsection{\textbf{Vector Field Based Trajectory Follower}}
Once the global trajectory is generated, we use a local vector field controller adapted from RAMP~\cite{2023ramp} to guide the robot towards the desired path while avoiding collisions and respecting biomechanical constraints. 
The controller shapes a directional field through attraction to the desired trajectory and repulsion from obstacles using negative gradients of the C-SDF, keeping the robot away from collisions while guiding it toward the goal.

At each time step $t$, the vector field controller outputs a candidate next configuration $\textbf{q}^R_{t+1}$. To maintain feasibility with respect to the human limb, we then apply the IK‐handshaking procedure (Algorithm~\ref{alg:handshake}) to ensure that $\textbf{q}^R_{t+1}$ remains within the coupled human–robot configuration space. 
If the handshaking step fails to find a valid pose for the arm (e.g., due to joint limits), we revert to $\textbf{q}^R_{t+1} = \textbf{q}^R_{t}$. That is, we pause the robot’s motion until a better command is generated. This safeguard prevents the system from taking any action that would force the human limb outside its feasible range.

Both the higher‐level trajectory generation and the local trajectory‐following loop run continuously until the robot reaches the final desired state $\textbf{q}^R_T$. Upon convergence, the resulting robot and human limb configurations are reported to any subsequent modules as shown in Fig. \ref{fig:pipeline} (e.g., a downstream application requiring the updated pose).

\subsection{Integration into a Bed‐Bathing Application}
We now demonstrate how our limb manipulation pipeline can be integrated into a bed‐bathing scenario, where one robot repositions the patient’s limb and another performs wiping. We define \emph{wiping targets} uniformly across the upper limb (e.g., near the upper arm and forearm) and reorder them to achieve a natural wiping sequence. Because we aim to clean multiple regions of the limb, rather than a single segment, the existing wiping module is not directly applicable. Thus, we introduce a selection scheme of \textit{feasible targets} to efficiently identify reachable wiping regions. 

Once the manipulation robot brings the limb to $\textbf{q}^R_T$ and $\textbf{q}^H_T$, it holds this configuration in place while the wiping robot cleans the designated targets. Since both robots operate in close proximity to the limb, collisions must be avoided. We therefore plan and execute the wiping robot’s motion using our method from Sec.~~\ref{sec:manipulation}, with the IK-handshaking step omitted, so the wiping robot can move freely to each target and then return to a resting pose without disturbing the limb-manipulation setup.

Before detailing our method for executing this bed bathing task, we first address how to select and manage suitable wiping targets and generate corresponding wiping trajectories. We also introduce a \textit{Next Arm Configuration Generator} to predict the next human limb configuration $\textbf{q}^H_{t+1}$ to which the arm should be repositioned for further cleaning. 

\subsubsection{\textbf{Selection of Feasible Targets}}
To check whether the wiping targets can be reached by the robot, we must perform IK. However, computing IK for every candidate point in a dense set of targets would incur a substantial computational burden, especially as the coverage area expands. In contrast, our feasible target selection method strategically prunes the search space through axis-wise partitioning (front and back segments) and by sampling a subset of representative points. This partitioning and sampling scheme significantly reduces the total number of IK queries, since only these representative target configurations undergo a full IK computation.

To formalize our target‐selection procedure, let 
\(\mathcal{V} = \{\mathbf{v}_1, \dots, \mathbf{v}_N\} \subset \mathbb{R}^3\) 
be the set of all available wiping targets. First, we choose an axis 
\(\mathbf{a}\in\mathbb{R}^3\) (e.g., the limb's ``front--back'' direction) and project each target \(\mathbf{v}_i\) onto this axis.
This yields scalar coordinates \(\alpha_i = \mathbf{a}^\top \mathbf{v}_i\). 
We then find the median \(\alpha_{\text{med}}\) and split the targets into two subsets:
\[
\mathcal{V}^{\text{front}} = \{\mathbf{v}_i \in \mathcal{V} \mid \mathbf{a}^\top \mathbf{v}_i 
    \ge \alpha_{\text{med}} \}, 
\quad
\mathcal{V}^{\text{back}} = \mathcal{V} \setminus \mathcal{V}^{\text{front}}.
\]
Next, we sample a smaller set of representative points 
\(\mathcal{S}^{\text{front}} \subseteq \mathcal{V}^{\text{front}}\) and 
\(\mathcal{S}^{\text{back}} \subseteq \mathcal{V}^{\text{back}}\). 
For each sampled point \(\mathbf{v}_i\), we solve inverse kinematics to check feasibility within the wiping robot's configuration space, including collision checks. 
If any \(\mathbf{v}_i\) in \(\mathcal{S}^{\text{front}}\) admits a feasible IK solution, 
the entire subset \(\mathcal{V}^{\text{front}}\) is considered potentially reachable (and likewise for \(\mathcal{V}^{\text{back}}\)). Otherwise, that subset is discarded to avoid 
unnecessary computations.

\subsubsection{\textbf{Wiping Trajectory Generation}}
Once a set of feasible wiping targets has been determined, we generate a wiping trajectory $\boldsymbol{\tau}^R_{wipe}$ that visits each reachable target in a logically ordered sequence. 
These targets are arranged in rows and reordered to follow a natural wiping pattern from one end of the arm to the other.
To ensure safety when the wiping robot moves closely around the manipulation robot and the human limb, we insert intermediate waypoints at the beginning and end of each segment. 
This creates a small safety distance that can reduce the risk of bumping into the human limb when the robot transitions between rows. 

Prior to generating the final trajectory, we also compute two additional transformations in the end effector frame: (1) $\textbf{T}_\text{target}^\text{eef}$ representing the target pose relative to the end effector, and (2) $\textbf{T}_\text{target\_iw}^\text{eef}$ representing the offset pose near the target relative to the end effector to facilitate a safer approach and departure. 
By solving inverse kinematics for these poses, we obtain corresponding joint angles $\boldsymbol{\theta}^R$ for the wiping robot. 
The final trajectory $\boldsymbol{\tau}^R_{\text{wipe}}$ is then built by appending valid target poses and their associated intermediate waypoints.

\subsubsection{\textbf{Next Limb Configuration Predictor}}
\label{sec:predictor}
Once the system completes a wiping motion, it may need a repositioning of the human limb to expose additional regions for cleaning. 
To achieve this, we train a neural network that takes into account the initial configuration of the human limb $\textbf{q}^H_{init}$ along with a \textit{labeled} point cloud of the limb $P^H_{labeled}$ and outputs an optimal goal $\textbf{q}^H_{goal}$. 

First, we obtain a point cloud $P^H$ of the limb and assign each point $\textbf{p}_i$ a label $\ell_i \in \{0, 1\}$ as follows:
\[
\ell_i \;=\;
\begin{cases}
1, & \text{if } \min_{\,v_j \,\in\, \mathcal{V}} \|\mathbf{p}_i - v_j\| \;\le\; r_{\mathrm{w}}, \\
0, & \text{otherwise}.
\end{cases}
\]
Here, $\mathcal{V}$ is the set of wiping targets, and $r_w$ is the radius that defines whether a point is \textit{within range} of a target. Intuitively, $\ell_i = 1$ indicates a region that requires further cleaning, and $\ell_i = 0$ indicates a region that may be irrelevant for the wiping task. This labeling provides the network with explicit cues about which parts of the limb still need attention.

To process this labeled point cloud, we use a PointNet-like encoder~\cite{qi2017pointnet, qi2017pointnetplusplus} that converts the set $\{(\textbf{p}_i, \ell_i)\}$ into a latent vector $\textbf{z}$, capturing geometric and task-relevant features from the limb surface.
Then, \textit{Next Limb Configuration Predictor} fuses this latent vector $\textbf{z}$ with the current limb configuration $\textbf{q}^H_{init}$ and predicts the goal configuration $\textbf{q}^H_{goal}$. 

We train this model in a supervised manner by collecting 2000$\times$14 data samples, each comrpising an initial configuration $\textbf{q}^H_{init}$, its labeled point cloud $P^H_{labeled}$, and an \textit{optimal} goal $\textbf{q}^H_{goal}$. For each of the 2000 initial limb configurations, we generate 14 variations of the wiping target distribution on the human limb. 
The optimal goal is found by sampling 100 candidate configurations per scenario and evaluating each with the scoring function as Eq.~\ref{eq:score} below.
\begin{equation}
\label{eq:score}
\begin{gathered}
    S_\text{feasibility} = \frac{1}{M} \sum_{i=0}^{M} r_i ; \quad 
    S_\text{closeness} = \bigl\| q_{H_{\text{init}}} - q_{H_{\text{goal}}} \bigr\|_2 ; \\
    S_\text{total} = w_\text{feasibility}*S_\text{feasibility} + w_\text{closeness}*S_\text{closeness} \\
\end{gathered}
\end{equation}
The configuration with the highest score is used as the training target. By unifying geometric labels from the point cloud with a task‐specific objective, the \textit{Next Limb Configuration Predictor} effectively proposes new limb poses that expose uncleaned regions for subsequent wiping.
\section{Evaluation}


\tikzset{
  labelfont/.style = {font=\scriptsize},
  myarrow/.style = {-{Stealth[length=1.2mm,width=1.2mm]}, line width=0.6pt}
}
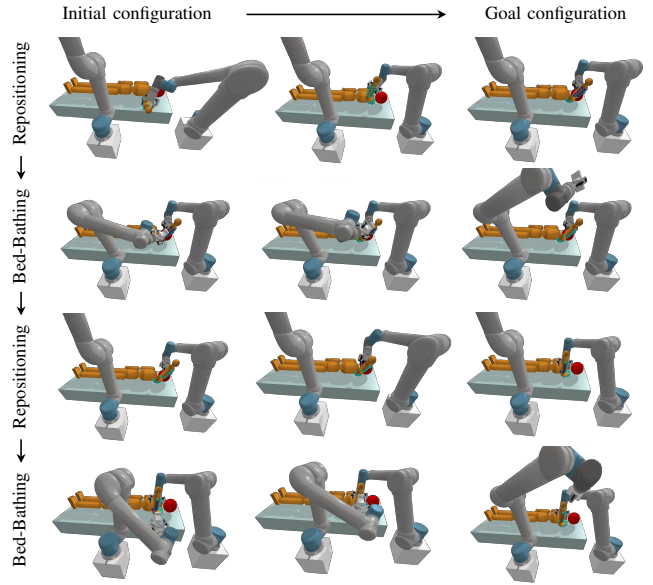
\begin{figure}[t]
\centering
\vspace{-1mm}
\begin{tikzpicture}[labelfont]
  \node[anchor=south west,inner sep=0] (img)
        at (0,0){\includegraphics[width=0.99\columnwidth,
                                   trim=0 5mm 0 0,clip]{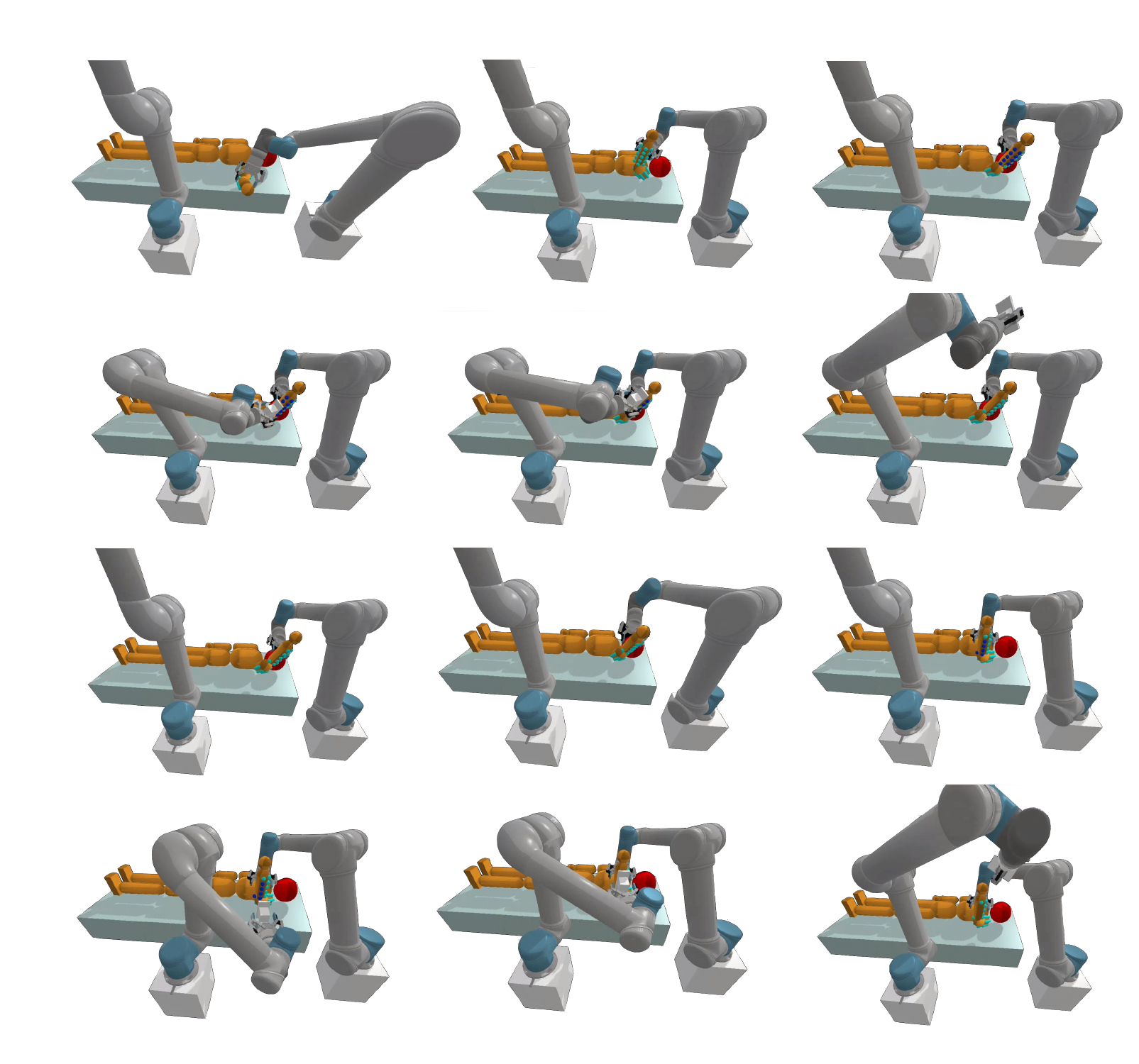}};
  \begin{scope}[x={(img.south east)}, y={(img.north west)}]

    \node[anchor=north] at (0.2,1.01) {Initial configuration};
    \node[anchor=north] at (0.85,1.01) {Goal configuration};
    \draw[myarrow] (0.37,0.98) -- (0.68,0.98);

    \node[anchor=east,rotate=90] at (0.02,0.95) {Repositioning};
    \draw[myarrow] (0.02,0.74) -- (0.02,0.7);

    \node[anchor=east,rotate=90] at (0.02,0.7) {Bed-Bathing};
    \draw[myarrow] (0.02,0.51) -- (0.02,0.47);

    \node[anchor=east,rotate=90] at (0.02,0.47) {Repositioning};
    \draw[myarrow] (0.02,0.26) -- (0.02,0.22);

    \node[anchor=east,rotate=90] at (0.02,0.22) {Bed-Bathing};

  \end{scope}
\end{tikzpicture}
\vspace{-4mm}
\caption{Image sequences for iteratively executing limb manipulation and bed-bathing task with a human in the supine position.}
\vspace{-3mm}
\label{fig:sim_envs}
\end{figure}

This section evaluates our limb manipulation system and its integration with a bed bathing application under various experimental conditions. We focus on the upper limb for our experiments and use a single robot arm to perform the manipulation under two scenarios: (1) a person lying in a supine position and (2) a person sitting upright. We also assess performance under different joint constraints that vary with different age groups. 
For integration of limb manipulation into bed bathing tasks, we use a dual-arm setup where one robot manipulates the human limb while the other performs bathing tasks, demonstrated with a human in supine position.
Fig.~\ref{fig:sim_envs} illustrates the experimental setup for the supine position.

\textbf{Metrics:} We consider the following metrics to evaluate the performance of our presented methods.
\textit{Success} denotes the number of trials in which the arm manipulation reaches its goal within one minute. We consider any trial exceeding this threshold as a failure. 
\textit{Total Time} measures the duration for the trajectory planner and follower to move the limb to its final configuration.
\textit{Plan Time} measures the time for our constrained MPPI planner to determine the motion trajectory.
\textit{Move Distance} measures the total Euclidean distance traveled by the limb in moving from the initial configuration to the desired configuration.
\textit{Out of Range} tracks joint-angle violations. 
We track 4-DOF arm angles \(\mathbf{q}^H = (q^H_0, q^H_1, q^H_2, q^H_3)\), representing shoulder flexion (yaw), shoulder external rotation (pitch), shoulder abduction (roll), and elbow angle, and increment this metric whenever a joint exceeds its feasible range. A value of zero indicates no violations throughout the motion.
For bed bathing, we measure \textit{Wiping Coverage} for the ratio of wiping points covered by the robot compared to the total target wiping points over the human limb,
\textit{Wiping Total Time} for planning and executing the wiping motion, 
\textit{Wiping Plan Time} for planning the wiping trajectory, and 
\textit{Wiping Travel Distance} for the total Euclidean distance traveled by the wiping robot to complete the wiping task.

\begin{figure}
    \centering
    \begin{subfigure}[b]{\columnwidth}
        \centering
        \includegraphics[width=0.4\textwidth]{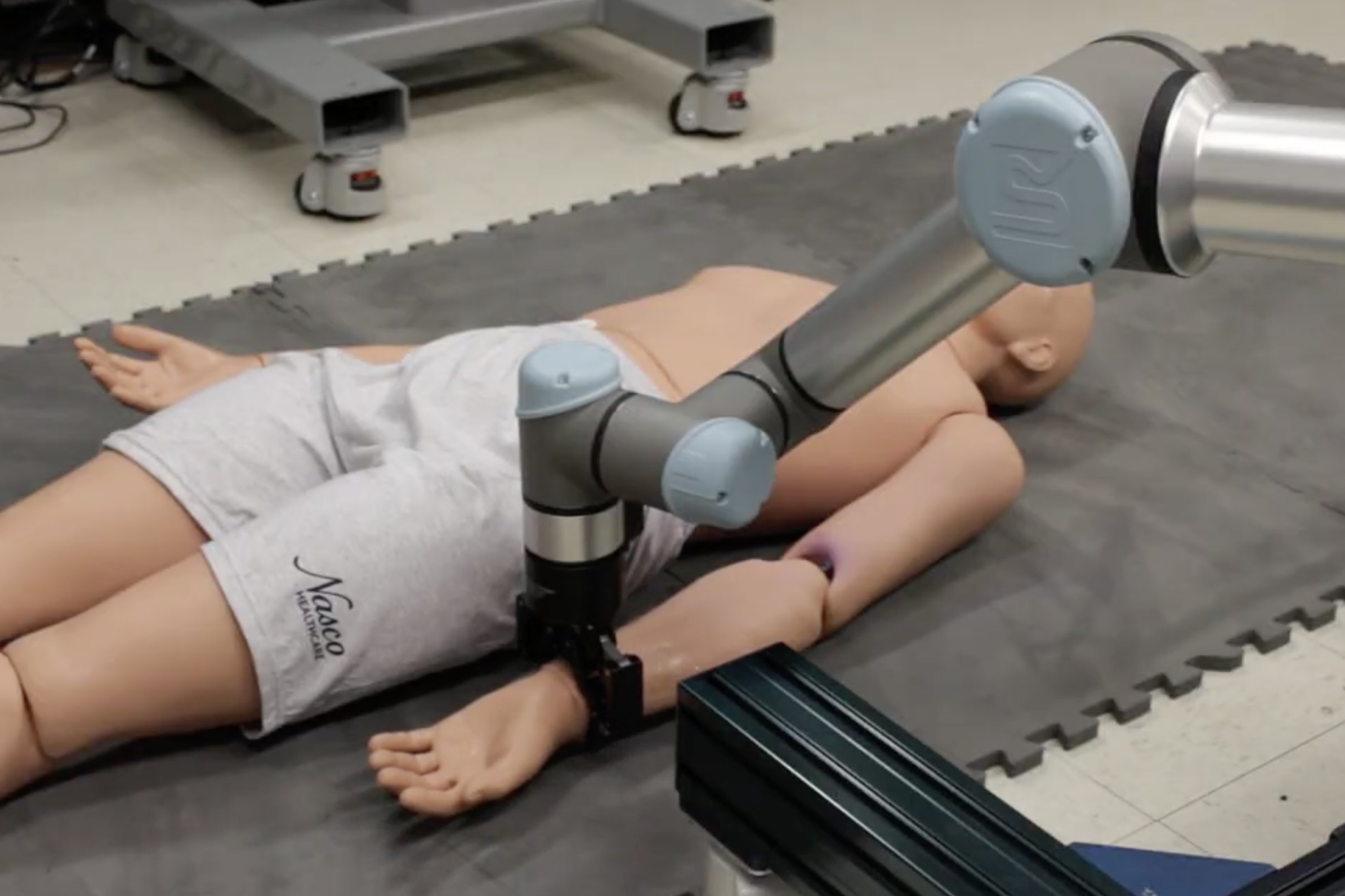}
        \includegraphics[width=0.4\textwidth]{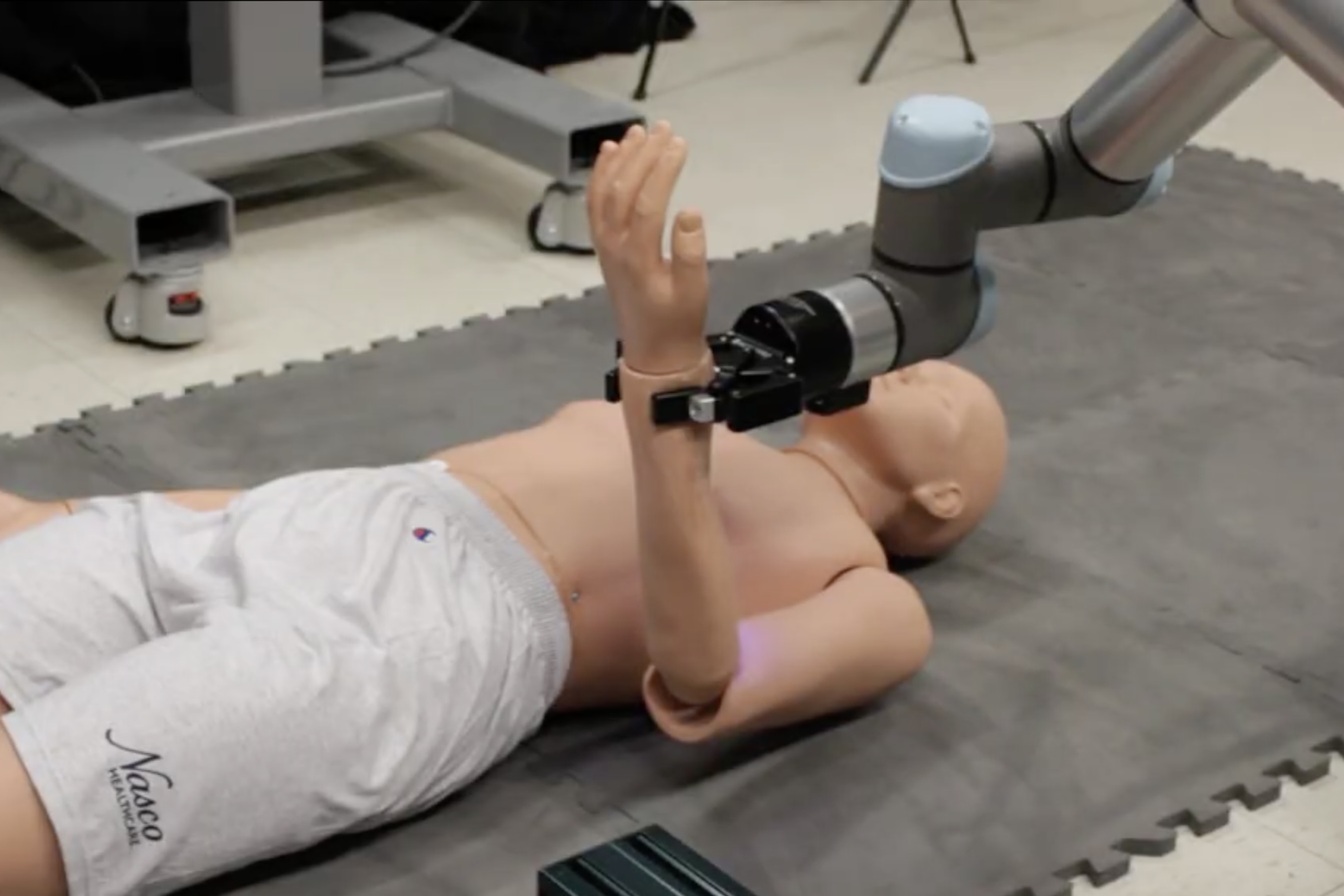}
    \end{subfigure}
    
    \vspace{0.4em} 

    \begin{subfigure}[b]{\columnwidth}
        \centering
        \includegraphics[width=0.4\textwidth]{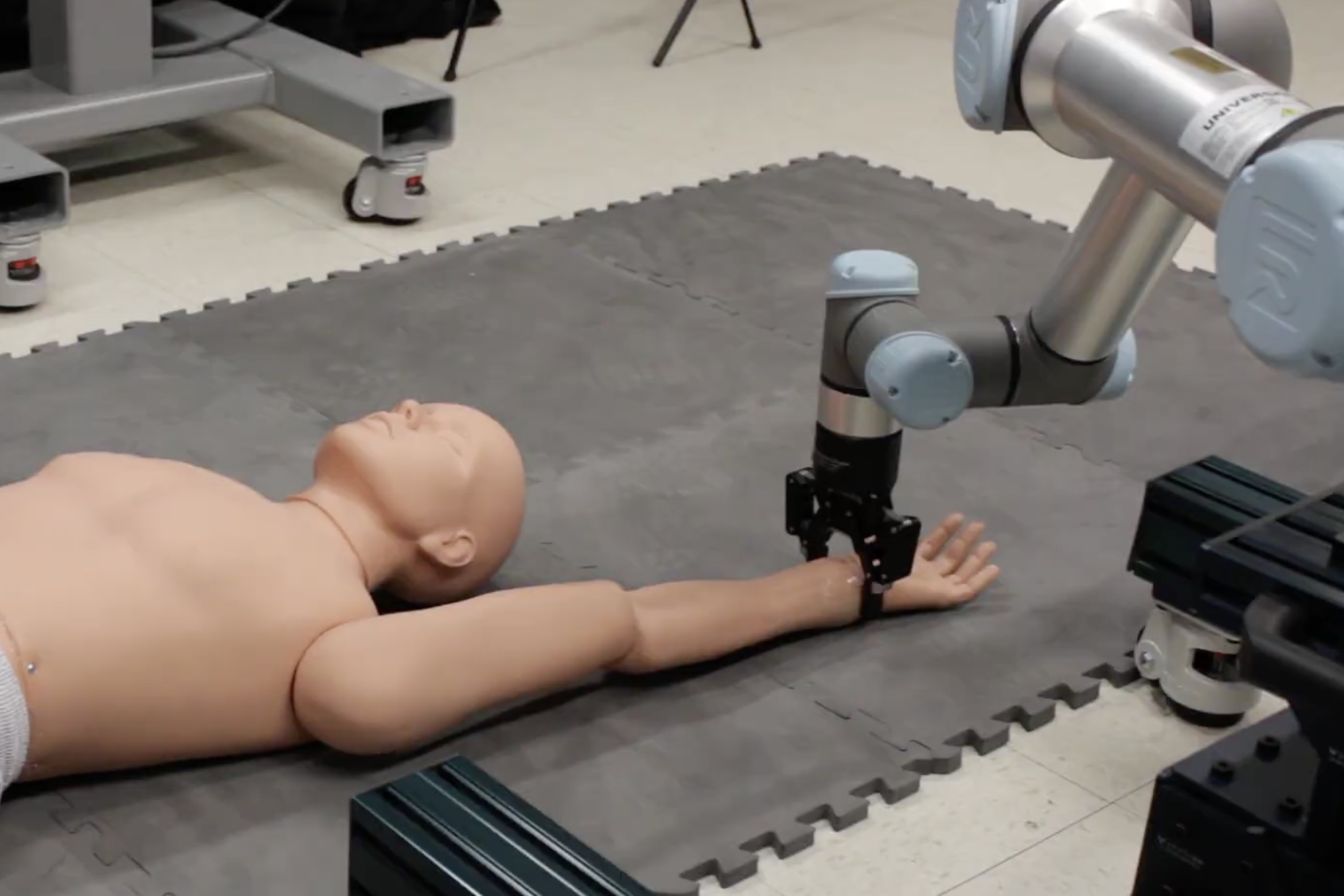}
        \includegraphics[width=0.4\textwidth]{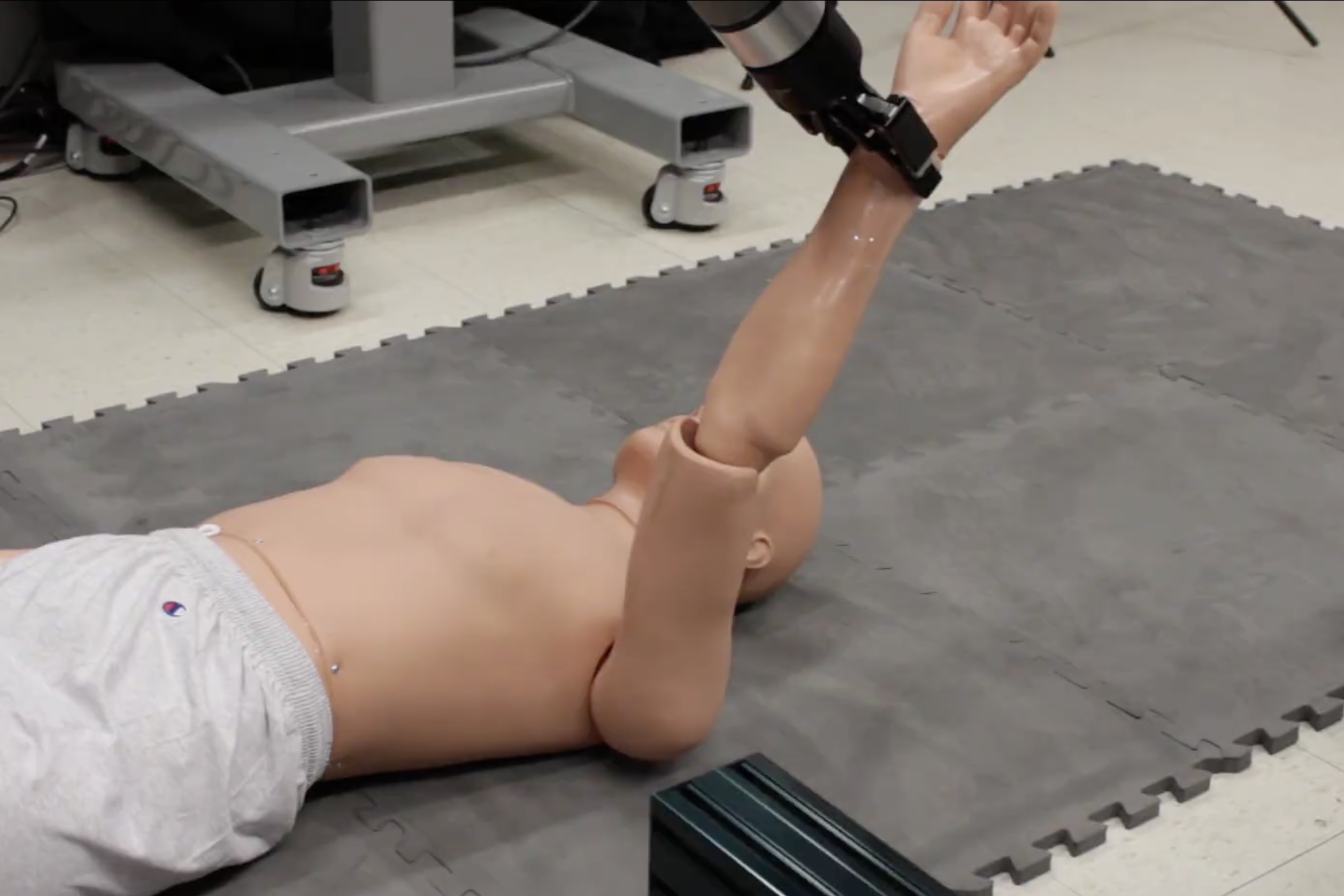}
    \end{subfigure}

    \caption{Real world demonstration of our approach featuring a human mannequin in the supine position, displaying different initial (left-most frame) and target (right-most frame) human arms configurations.}
    \vspace{-3mm}
    \label{fig:real_exps}
\end{figure}

\textbf{Evaluating Limb Manipulation:}
Since our limb manipulation approach is the first of its kind to address these tasks, we do not have comparable baselines to reference. Instead, we evaluate the performance of our approach across a diverse range of limb configurations and varying biomechanical constraints.
Additionally, one might consider using Sampling-based Motion Planners (SMPs) to generate motion planning trajectories that adhere to collision avoidance and closed-loop biomechanical  constraints \cite{kingston2018ar}. While theoretically viable, SMPs are often too slow in closed-loop tasks, making them impractical. Additionally, we observed that trajectory following often requires replanning of limb manipulation trajectories in real time, complicating SMP integration.
One might also consider replacing our velocity-field follower with model predictive control (MPC) \cite{2016mppi, kemp2016reposition}. However, MPC must solve a constrained optimization at every step, leading to latency and abrupt command changes. In comparison, our method simply projects each candidate update onto the coupled human–robot manifold with a single analytic IK call, making it more efficient. Moreover, obstacle avoidance is driven directly by the gradient of the C-SDF, so the controller reacts immediately to newly perceived obstacles \cite{2023ramp}.

\textbf{Evaluating Bed Bathing:}
In our evaluation, we integrate limb manipulation into bed bathing by generating a sequence of target limb poses to maximize wiping coverage. We compare a sampling-based approach with a neural network-based generator and also assess the improvement over a static human pose. We highlight the benefits of limb manipulation compared to a static human pose for conducting wiping task.

Finally, all experiments are conducted using UR5 robots. The simulation was established within the PyBullet simulation environment~\cite{coumans2021pybullet}. We use a
system with a 3.50GHz $\times$ 8 Intel Core i9 processor, 32 GB RAM, and GeForce RTX 3090 GPU for all processing and computations.


\subsection{Limb Manipulation}

We first evaluate how effectively our method manipulates a human arm into various goal configurations while respecting joint limits. In each setting (supine or sitting posture), we execute our limb manipulation procedure using two different grasps. For each grasp, we sample 5 random initial arm configurations and 10 random goal configurations, for a total of $2\times5\times10 = 100$ trials per posture.

To further explore the variability introduced by different mobility and flexibility needs, we incorporate shoulder range reductions reported by~\cite{gill2020shoulder}. Specifically, we reduce the shoulder’s maximum flexion, external rotation, and abduction angles by the age groups as follows: 
\begin{itemize} 
    \item \textbf{20--39 years:} (0.2617, 0.3086, 0.4278) rad
    \item \textbf{40--59 years:} (0.3544, 0.5720, 0.5149) rad
    \item \textbf{60--79 years:} (0.5734, 0.7114, 0.7617) rad
    \item \textbf{80+ years:} (0.8175, 0.7240, 1.0662) rad
\end{itemize} 
Note that each age group combines the results from both male and female participants. 

\begin{table}[ht]
    \centering
    \caption{Arm Manipulation Evaluation in Supine and Sitting Positions in 100 Trials}
    \begin{tabular}{c|c|c|c}
        \toprule
        \multicolumn{2}{c}{} & Supine Position & Sitting Position \\
        \midrule
        \multicolumn{2}{c|}{Success} & 97/100 & 92/100 \\
        \midrule
        \multirow{4}{5em}{\centering Out of Range (rad)} 
        & $q_0$ & 0.0 $\pm$ 0.0 & 0.0 $\pm$ 0.0 \\
        & $q_1$ & 0.0001 $\pm$ 0.001 & 0.0 $\pm$ 0.0\\
        & $q_2$ & 0.0 $\pm$ 0.0 & 0.0 $\pm$ 0.0\\
        & $q_3$ & 0.0 $\pm$ 0.0 & 0.0 $\pm$ 0.0003 \\
        \midrule
        \multicolumn{2}{c|}{Total Time (s)} & 7.3866 $\pm$ 6.6760 & 10.0573 $\pm$ 6.6397\\
        \midrule
        \multicolumn{2}{c|}{Plan Time (s)} & 4.1290 $\pm$ 3.6406 & 3.7494 $\pm$ 3.4953\\
        \midrule
        \multicolumn{2}{c|}{Move Dist. (m)} & 0.1861 $\pm$ 0.1371 & 0.2689 $\pm$ 0.1705\\
        \bottomrule
    \end{tabular}
\label{tab:results_manip}
\end{table}

\begin{table*}[ht]
    \centering
    \caption{Arm Manipulation Evaluation in Supine and Sitting Position with Different Shoulder Ranges in 100 Trials (values in paranthesis indicate the maximum joint angle that was out of range)}
    \begin{tabular}{cc|c|c|c|c|c}
        \toprule
        \multicolumn{3}{c}{} & \multicolumn{4}{c}{{Age Group by Years}} \\
        \multicolumn{3}{c}{} & {20-39} & {40-59} & {60-79} & {80+} \\
        \midrule
        
        \multirow{8}{*}{\centering {Supine}}
            & \multicolumn{2}{c|}{{Success}} & 99/100 & 97/100 & 99/100 & 98/100 \\
            \cmidrule{2-7}
            & \multirow{4}{5em}{\centering {Out of Range (rad)}} 
                & $q_0$ & 0.0 $\pm$ 0.0 & 0.0 $\pm$ 0.0 & 0.0005 $\pm$ 0.005 (0.066) & 0.0 $\pm$ 0.0 \\
                & & $q_1$ & 0.0 $\pm$ 0.0001 (0.0003) & 0.0 $\pm$ 0.0 & 0.0 $\pm$ 0.0 & 0.001 $\pm$ 0.011 (0.134) \\
                & & $q_2$ & 0.0 $\pm$ 0.0001 (0.004) & 0.0 $\pm$ 0.0 & 0.0 $\pm$ 0.0 & 0.0 $\pm$ 0.0 \\
                & & $q_3$ & 0.0 $\pm$ 0.0 & 0.002 $\pm$ 0.017 (0.176) & 0.0 $\pm$ 0.0 &  0.0 $\pm$ 0.0 \\
            \cmidrule{2-7}
            & \multicolumn{2}{c|}{{Total Time (s)}} & 7.5756 $\pm$ 4.9940 & 5.9115 $\pm$ 4.6720 & 5.1445 $\pm$ 3.5540 & 6.4180 $\pm$ 7.0860\\
            \cmidrule{2-7}
            & \multicolumn{2}{c|}{{Plan Time (s)}} & 4.3838 $\pm$ 3.5680 & 3.1211 $\pm$ 2.6642 & 2.7460 $\pm$ 1.8543 & 3.1880 $\pm$ 3.2017 \\
            \cmidrule{2-7}
            & \multicolumn{2}{c|}{{Move Dist. (m)}} & 0.1721 $\pm$ 0.1131 & 0.1456 $\pm$ 0.1238 & 0.1212 $\pm$ 0.1437 & 0.1400 $\pm$ 0.1878 \\
        \midrule
        
        \multirow{8}{*}{\centering {Sitting}}
            & \multicolumn{2}{c|}{{Success}} & 94/100 & 91/100 & 96/100 & 91/100 \\
            \cmidrule{2-7}
            & \multirow{4}{5em}{\centering {Out of Range (rad)}} 
                & $q_0$ & 0.0 $\pm$ 0.0 & 0.0 $\pm$ 0.0 & 0.0006 $\pm$ 0.008 (0.127) & 0.0001 $\pm$ 0.001 (0.039) \\
                & & $q_1$ & 0.0 $\pm$ 0.0004 (0.017) & 0.002 $\pm$ 0.012 (0.170) & 0.008 $\pm$ 0.041 (0.495) & 0.003 $\pm$ 0.015 (0.146) \\
                & & $q_2$ &  0.0 $\pm$ 0.0 & 0.0 $\pm$ 0.0 & 0.0 $\pm$ 0.0 & 0.0 $\pm$ 0.0004 (0.014) \\
                & & $q_3$ & 0.0 $\pm$ 0.0004 (0.018) & 0.0 $\pm$ 0.0001 (0.011) & 0.0004 $\pm$ 0.004 (0.057) & 0.0001 $\pm$ 0.0006 (0.018) \\
            \cmidrule{2-7}
            & \multicolumn{2}{c|}{{Total Time (s)}} & 8.9741 $\pm$ 6.4800 & 8.9121 $\pm$ 8.5780 & 8.1351 $\pm$ 5.9398 & 7.7969 $\pm$ 6.8693 \\
            \cmidrule{2-7}
            & \multicolumn{2}{c|}{{Plan Time (s)}} & 3.7842 $\pm$ 3.1794 & 4.6091 $\pm$ 4.7414 & 3.5151 $\pm$ 3.1310 & 3.6791 $\pm$ 3.7322 \\
            \cmidrule{2-7}
            & \multicolumn{2}{c|}{{Move Dist. (m)}} & 0.2460 $\pm$ 0.1548 & 0.2350 $\pm$ 0.1727 & 0.2271 $\pm$ 0.1535 & 0.2216 $\pm$ 0.1540 \\
        \bottomrule
    \end{tabular}
\label{tab:results_manip_group_supine_sitting}
\vspace{-0mm}
\end{table*}

Table~\ref{tab:results_manip} shows consistently high success rates over 90\% for both supine and sitting positions, with minimal joint-limit violations across all four degrees of freedom. Most out-of-range values remain close to zero, indicating that the robot successfully avoids exceeding biomechanical constraints during motion. Table~\ref{tab:results_manip_group_supine_sitting} demonstrates robustness across different shoulder mobility ranges, maintaining success rates above 90\% even when shoulder flexion, external rotation, or abduction are restricted in older age groups.
The short manipulation and planning times, combined with modest travel distances, further validate our approach's viability for real-world assistive tasks.

We also evaluate our limb manipulation approach with a real-world human mannequin (Fig.~\ref{fig:real_exps}). 
In these trials, a UR5 robot repositions the mannequin's arm from two different initial configurations to target goal configurations.
Because the mannequin's shoulder is limited to a single degree of freedom, unlike a real human shoulder with three degrees of freedom, its kinematic model and joint ranges differ from those used in simulation. Despite this simplification, the experiments demonstrate successful limb repositioning, suggesting robustness to variations in arm geometry.


\subsection{Integrated Pipeline with Bed Bathing Tasks}

\begin{table}[ht]
\vspace{-0mm}
    \centering
    \caption{Bed Bathing Task Evaluation in a Supine Position with a Static Human vs. Repositioning in 100 Trials}
    \begin{tabular}{cc|c}
        \toprule
        & \multicolumn{2}{c}{\text{Wiping Coverage}} \\
        \midrule
        \multirow{2}{*}{\text{Static Human}} & all trials & 0.053 $\pm$ 0.087 \\
        & succ. trials & 0.142 $\pm$ 0.086\\
        \midrule
        \text{Limb Manip \& NN Config Pred} & all trials & 0.497 $\pm$ 0.064 \\
        \midrule
        \text{Limb Manip \& Rand Config Pred} & all trials & \textbf{0.564 $\pm$ 0.106} \\
        \bottomrule
    \end{tabular}
\label{tab:results_wiping_static}
\vspace{-0mm}
\end{table}

\begin{table}[ht]
    \centering
    \caption{Arm Manipulation and Bed Bathing Task Evaluation in Supine Position in 100 Trials}
    \begin{tabular}{c|c|c|c}
        \toprule
        \multicolumn{2}{c|}{} & \text{Use Next Arm} & \text{Use Random}  \\ 
        \multicolumn{2}{c|}{} & \text{Predictor} & \text{Generator} \\ 
        \midrule
        \multicolumn{2}{c|}{\text{Success}} & \textbf{91/100} & 73/100 \\
        \midrule
        \multicolumn{2}{c|}{\text{Total Time (s)}} & \textbf{140.03 $\pm$ 26.240} & 699.10 $\pm$ 63.90 \\
        \midrule
        \multirow{3}{4em}{\centering {Arm Manip.}} 
        & Total Time (s) & \textbf{4.7433 $\pm$ 2.490} & 7.760 $\pm$ 3.740 \\
        & Plan Time (s) & \textbf{2.4884 $\pm$ 1.360} & 3.030 $\pm$ 1.430 \\
        & Move Dist. (m) & \textbf{0.7914 $\pm$ 0.240} & 1.860 $\pm$ 0.330 \\
        \midrule
        \multirow{4}{4em}{\centering {Bed Bathing (Wiping)}} 
        & Coverage & 0.4974 $\pm$ 0.0641 & \textbf{0.5642 $\pm$ 0.1064} \\
        & Total Time (s) & \textbf{7.5826 $\pm$ 9.010} & 8.590 $\pm$ 10.250 \\
        & Plan Time (s) & \textbf{0.1611 $\pm$ 0.0950} & 0.1690 $\pm$ 0.1010 \\
        & Travel Dist. (m) & \textbf{14.0732 $\pm$ 2.450} & 14.790 $\pm$ 3.520 \\
        \midrule
        \multicolumn{2}{c|}{\text{Next Goal Gen. Time (s)}} & \textbf{0.8967 $\pm$ 0.1080} & 52.680 $\pm$ 10.910 \\
        \bottomrule
    \end{tabular}
\label{tab:results_integrated}
\vspace{-2mm}
\end{table}

To demonstrate our limb manipulation system in a complete assistive scenario, we integrate it into a bed bathing pipeline using a single specified grasp pose with the human in the supine position. 
During up to 10 iterations, the system repositions the human arm and conducts a wiping task, recording performance metrics after each iteration. We repeat this setup for a total of 100 trials. Fig. \ref{fig:sim_envs} demonstrates a fragment of one trial of integrated bed bathing and limb manipulation using our approach. To explore different decision-making strategies to select the subsequent configuration of the human arm, we compare two approaches: (1) a neural network model that predicts the optimal next arm configuration (Sec.~\ref{sec:predictor}), and (2) a random sampling approach that generates 50 candidate configurations and selects the one with the highest evaluated score (Eq.~\ref{eq:score}).

Table~\ref{tab:results_wiping_static} first compares a static-arm scenario against one in which the human arm is actively repositioned. In the static case, we evaluate with 10 different wiping robot base poses and 10 different human arm configurations, where the wiping robot is placed near the human arm to improve reachability. However, some robot–arm configurations still fail to reach the wiping target, resulting in a low average coverage. When considering only those trials that achieve nonzero coverage, the mean coverage value is higher but remains significantly lower than what is achieved by repositioning the arm. In contrast, actively moving the limb significantly increases average wiping coverage, highlighting the benefit of repositioning to expose more cleaning surfaces.

Table~\ref{tab:results_integrated} further examines our integrated bed bathing pipeline.
We define \textit{task success} as completing both arm manipulation and wiping robot movement to designated targets within one minute. Trials exceeding this threshold are considered failures. 
\textit{Total time} measures the duration of each trial.
Both neural network and random sampling approaches yield substantially higher wiping coverage than the static-arm baseline. 
However, the neural network approach shows greater overall efficiency with higher success rates and significantly shorter total execution times.
While random sampling method yields slightly higher average coverage, it incurs much larger planning overhead, making it less practical for time-sensitive assistive tasks. 
The neural network method's lower coverage, however, suggests it may occasionally miss arm configurations that yield optimal coverage.


\section{Conclusions and Future Work}
In this work, we present a unified simulation framework for robotic limb manipulation with tools ranging from grasping to trajectory generation and following. We show that our pipeline robustly maintains biologically plausible joint constraints across various poses and limb configurations. We further demonstrate that integrating limb manipulation with a bed bathing system significantly improves assistive task performance. Overall, these findings suggest that strategic limb repositioning can expand the capabilities of assistive robots in healthcare settings.

Our proposed work serves as an initial platform and reference implementation for limb manipulation research, laying the groundwork for future developments. Currently, we assume a passive human model, but we plan to integrate active human dynamics to reflect the implicit joint torques present in real humans. Additionally, we represent the human body as rigid links, whereas a more realistic model would include rigid bones and soft skin. We aim to enhance our simulator to incorporate this soft-rigid body representation in future work.



\section*{ACKNOWLEDGMENT}
We thank Jainam Doshi for his help on implementing the human URDF model in the sitting position.



\bibliographystyle{IEEEtran}
\bibliography{references}

\end{document}